# A Comparison of Document Similarity Algorithms

Nicholas Gahman and Vinayak Elangovan

Computer Science Program, Division of Science and Engineering,
Penn State University at Abington, Pennsylvania, USA

## *Abstract*

Document similarity is an important part of Natural Language Processing and is most commonly used for plagiarism-detection and text summarization. Thus, finding the overall most effective document similarity algorithm could have a major positive impact on the field of Natural Language Processing. This report sets out to examine the numerous document similarity algorithms, and determine which ones are the most useful. It addresses the most effective document similarity algorithm by categorizing them into 3 types of document similarity algorithms: statistical algorithms, neural networks, and corpus/knowledge-based algorithms. The most effective algorithms in each category are also compared in our work using a series of benchmark datasets and evaluations that test every possible area that each algorithm could be used in.

## 1. Introduction

Document similarity analysis is a Natural Language Processing (NLP) task where two or more documents are analyzed to recognize the similarities between these documents. Document similarity is heavily used in text summarization, recommender systems, plagiarism-detection as well as in search engines. Identifying the level of similarity or dissimilarity between two or more documents based on their content is the main objective of document similarity analysis. Commonly used techniques like cosine similarity, Euclidean distance, etc., compares the document's text features to provide a similarity score between 0 and 1, where 1 indicates complete similarity and 0 indicates no similarity. Although there are numerous algorithms used for document similarity, there are no algorithms universally recognized as the most effective or efficient in a given area.

This paper seeks to rectify this issue by categorizing the document similarity algorithms into 3 different types of document similarity algorithms: statistical algorithms, neural networks, and corpus/knowledge-based algorithms. Each category's algorithms will be closely inspected, and several algorithms from each category will be judged to be the best of said category. Once the most effective algorithms of each category are found, they will be run through four different datasets and three different metrics. Finally, once this data is obtained, it will be analyzed to determine which of the algorithms are most effective.

This paper is organized as follows: In section 2: Relevant Works, a variety of similar research work are discussed. In section 3: Proposed Methodology: a general description of three categories of document similarity algorithms is given, and the methodology and the comparison process are more thoroughly discussed. Finally, in section 4: Results, the specifics, and limitations of implementation of each algorithm are discussed, the results of the comparison process are given followed by conclusions in section 5.

## 2. RELEVANT WORKS

Jesus M. Sanchez-Gomez, along with other researchers, compared term-weighting schemes and similarity measures when used for extractive multi-document text summarization [1]. The result of this research is that overall cosine similarity appears to be the most effective statistical similarity measure in this area. Kazuhiro Seki created a text similarity measure that was able to work across languages [2]. The proposal accomplished this by using two Neural Machine Translation models to build word embeddings that it can use to compute document similarity. The NMTs in question translated multiple possible translations to account for mistranslations. The result of the translations are matrices that are normalized and transformed, creating multilingual word embeddings that can be compared using cosine similarity. The proposal has significant flexibility, as the study explicitly states that it is possible to add other similarity scores onto this system. This proposal was compared to other multilingual similarity algorithms such as Doc2Vec, Sec2Vec, and the S2Net Siamese Neural Network, with the result being that it outperformed all other algorithms. When the proposal's sentence retrieval was compared to Google Translate, they had similar results, despite the proposal having a much lower BLEU score due to its relatively small training data and models.

Emrah Inam proposed a document similarity algorithm that combines the usage of word embeddings with knowledge-based measures through a semantic lexicon named ConceptNet [3]. Once the sentences are run through ConceptNet, the proposal then produces a vector representation of the transformed sentences using a breadth first traversal algorithm. Finally, a soft cosine similarity measure is used to compare the vectorized sentences. The similarity from the ConceptNet metric and the similarity from the dependency parser model is then combined to produce the final similarity score. The proposal, named SimiT, is given the Microsoft Research Paraphrase Corpus (MRPC) as an evaluation task to determine if it can detect and understand paraphrases. Its result is measured using Pearson correlation and is then compared to several other similarity algorithms' results. Among these algorithms are basic cosine similarity, Word2Vec cosine and soft cosine, and several state-of-the-art methods such as BERT and its variations. Of the compared algorithms, SimiT performed very well, exceeding the performance of all but the state-of-the-art methods. Even among the state-of-the-art methods, SimiT was still useful due to its incredibly low run time.

Table 1. Chosen dataset's tasks and purpose.

|  | **MRPC** | **AFS** | **SICK-R** | **SICK-E** |
|---|---|---|---|---|
| **Benchmark Purpose** | Testing if algorithm can detect paraphrases | Testing if algorithm can detect similar arguments | Testing if algorithm can see lexical similarity of sentences | Testing if algorithm can see semantic similarities/differences between sentences |

Aminul Islam and Diana Inkpen collaborated on a method to give knowledge-based measures the ability to determine sentence similarity [4]. Knowledge-based measures are designed primarily to compare concepts, and while turning a word into a concept is easy it is much harder to do so for a sentence. As part of the process, the authors introduce a string similarity word method using three modifications of the longest common subsequence: The Normalized Longest

Common Subsequence (NLCS), the Normalized Maximal Consecutive Longest Common Subsequence starting from the first character (NMCLCS$_1$), and the Normalized Maximal Consecutive Longest Common Subsequence starting from any character (NMCLCS$_n$). The scores from each of the three modified LCS algorithms are combined, then divided by three, such that there is an average between 0 and 1.

To develop a sentence similarity algorithm that uses knowledge-based measures, the authors first process the sentences by removing all special characters, punctuation, capital letters, and stop words. At this point, the lengths of the two sentences are stored for later use. From here, they then find all of the words in both sentences that match with each other using the knowledge-based measures or string similarity measures, count the number of words where this is the case, and remove all matching words from the sentence. Then, all the remaining words are compared using the knowledge-based measure, and the results are put in a matrix whose length is equal to the length of the first sentence and width is equal to the length of the second sentence. Another matrix is created using the same method, with the string similarity method replacing the knowledge-based measure. The knowledge-based measure matrix is added with the string similarity matrix to produce a joint matrix. Next, the highest value in the joint matrix is found and added to a list, after which the column and row the value was in is deleted. This continues until the highest value is zero or until there are no rows or columns left. The final algorithm to determine the similarity of the two sentences is the number of matching words plus all the values in the list, times the length of the two sentences without stop words combined. The result is then divided by two times the length of the first sentence without stop words, times the length of the second sentence without stop words. Table 2 summarizes the advantages and disadvantages of different algorithm categories.

Table 2. Algorithm categories.

|  | **Advantages** | **Disadvantages** |
|---|---|---|
| **Basic Statistical Techniques** | Simple, easy to use | Does not obtain enough semantic information to make accurate predictions about entailment vs contradiction |
| **Neural Networks** | Incredibly effective, achieves state-of-the-art results | Incredibly computationally expensive and memory intensive, difficult to debug |
| **Knowledge/Corpus-Based Measures** | Can trivially obtain and process the semantic information needed for accurate predictions | Relies heavily on large corpora and semantic networks to work properly, is designed for word/concept similarity and thus difficult to scale |

Yigit Sever and GonencErcan perform a comparison of cross-lingual semantic similarity methods by building a cross-lingual textual similarity evaluation dataset utilizing seven different languages [5]. Wordnets measure concepts, and while some concepts are specific to a given language, others are shared between languages. By linking several of these synsets together, a foundation is built for building the evaluation dataset. The resulting dataset can determine the effectiveness of both unsupervised and supervised text similarity algorithms, allowing the two categories to be easily compared. Word embeddings are victim to the hubness problem, where many of the vector embeddings are similar to other vector embeddings. To investigate how hubs such as these can affect the similarity result, the evaluation task is split into two different tasks: alignment and pseudo-retrieval.

In addition to building the evaluation dataset, the paper also compares several text similarity models, both unsupervised and supervised. Among the algorithms tested is machine translation the monolingual baseline (MT), cosine similarity between sentence embeddings (SEMB), word mover's distance (WMD), Winkhorn (SNK), and Siamese long-short term memory (LSTM). Concerning retrieval, WMD surprisingly performed the best, followed by SNK. LSTM performed the third best, but it was noted that it was trained using a limited number of instances, so it may be possible for it to score higher in this area. Concerning the matching task, SNK performed the best, followed by WMD, followed by SEMB.

Yuki Arase and Junichi Tsuji developed a method of improving the BERT model through transfer fine tuning [6]. Their method of pretraining focused on semantic pair modeling, allowing the proposed method to have significant improvements over the normal BERT model. Furthermore, in addition to performing better than the baseline BERT model, it is also more cost-effective, as it focuses on phrase alignments, which can be automatically generated.
As part of the evaluation process, the paper used two different benchmarks: the GLUE Benchmark, and the PAWS dataset. The GLUE Benchmark used consists of nine different evaluation datasets whose tasks cover various parts of natural language understanding. The most relevant tasks for the paper, however, is that of Semantic Equivalence, whose subtasks consist of paraphrase understanding and an understanding of Semantic Text Similarity (STS). The PAWS dataset, like the MRPC dataset, focuses on paraphrase understanding, but utilizes controlled word swapping and back translation to determine if an algorithm is sensitive to context and word order. In the GLUE Benchmark, the proposed method successfully performed better than their BERT counterpart in all areas except for QNLI and SST, while in the PAWS benchmark the proposed method performed better than BERT in all areas. Furthermore, the evaluation showed that the proposed method performed better when the fine-tuned training corpus used is smaller, which would make creating said training corpi much easier.

## 3. PROPOSED METHODOLOGY

Statistical techniques are the simplest of the three types of document similarity algorithms. They compare text by first turning the sentences into vectors, and then comparing said vectors. The most used and most effective way of comparison is through cosine similarity, but other methods such as Euclidean distance are occasionally used. Of the possible preprocessors, Sentence-BERT was judged to be most ideal for the purposes of this paper. This is because it performed incredibly well among the algorithms tested for this purpose and was the second most computationally efficient algorithm tested [7]. Neural networks are another possible avenue of document similarity, and a very effective one. The basics of neural network-based techniques is that they are first fed training documents of pairs of texts that are either similar to each other or different. Gradually, the neural network learns which pairs of texts are similar and which pairs are different through understanding semantic information. Generally, the text is run through a tokenizer first, which allows the semantic information to be more understandable to the network. Some models, like BERT and XLNet, are pre-trained, which allows the network to have accurate predictions without needing long training times and massive corpora for each individual task. Figure 1 shows a typical methodology of document analysis using statistical algorithms.

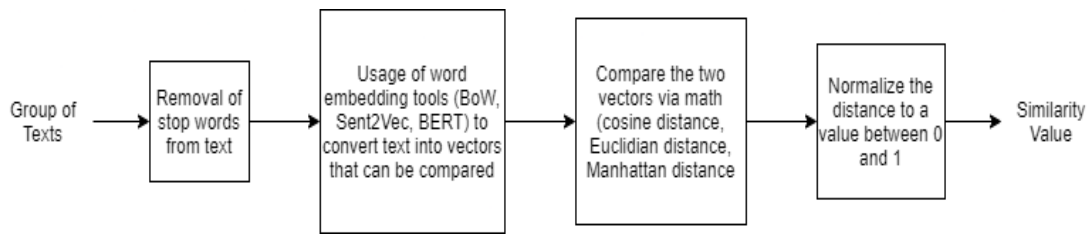

Figure 1. Methodology of statistical algorithms.

Corpus-based measures find text similarity by using large corpora to determine the similarity between different words [8]. Two methods that utilize this technique are Pointwise Mutual Information, and Latent Semantic Analysis. Pointwise Mutual Information accomplishes this by determining how often the compared words appear together, while Latent Semantic Analysis accomplishes this by performing singular value decomposition on the corpus. Knowledge-based measures follow a similar path by converting words into concepts, and then using semantic networks to compare those concepts [8].

Two neural-network systems are included, as they seem to be overall most effective. These are knowledge-distilled MT-DNN, and XLNet [9]. The Multi-Task Deep Neural Network proposed in [10] is improved using knowledge distillation. Several "teacher" MT-DNNs are developed, one for each task. Each "teacher" MT-DNN then generates a set of "soft" targets, which are combined with the correct "hard" targets for their respective task. A single "student" MT-DNN then uses these combined targets as the target for each task. The end result of this process is that the "student" MT-DNN is able to significantly outperform other MT-DNNs that do not use knowledge distillation as part of their training. XLNet combines the usage of both autoregressive language modeling and autoencoding to build a pretrained model that can obtain context from both previous and future words without relying on data corruption like BERT [11]. As a result, XLNet is able to significantly outperform the baseline BERT in most NLP tasks. In addition, a statistical technique is also included, dubbed Sentence-BERT + cosine similarity. Sentence-BERT is a modification of BERT that uses siamese triplet networks to derive sentence embeddings that are both semantically meaningful and can easily be compared using cosine similarity [7]. In addition to these algorithms, a combination of the Lin knowledge-based measure proposed in [8] and a string similarity method proposed in [12] and [4] is also used. The Lin knowledge-based measure is designed specifically to understand semantic similarity, while the string similarity method was designed to understand lexical similarity. As a result, the combination of the two algorithms would be able to understand far more information than each of the algorithms individually.

The datasets used will be the Microsoft Research Paraphrase Corpus (MRPC) [13], the Argument Facet Similarity (AFS) dataset [14], and the Sentences Involving Composition Knowledge (SICK) dataset [15]. The MRPC dataset is designed to test whether the algorithm can detect paraphrases, while the AFS dataset is designed to test whether the algorithm can detect similar arguments. The SICK dataset is split into two tasks, named SICK-R and SICK-E. SICK-R tests the algorithm on the sentences' lexical similarity, while SICK-E tests whether the algorithm is capable of high-level semantic knowledge, by deciding whether the sentences are similar, have no relation, or have the opposite meaning. The evaluations used for the tasks will be the Pearson correlation and Spearman rank correlation for AFS and SICK-R, and classification accuracy for MRPC and SICK-E.

## 4. RESULTS

Overall, the implementation of Sentence-BERT + cosine similarity went well. For the most part, the corpuses, once downloaded, were trivial to implement. The biggest issue came from the AFS dataset, as it was stored in a comma separated value format, which made it difficult to distinguish the commas in sentences from the commas separating the data. This issue was solved by editing the dataset to make every comma in a sentence appear twice, and then use regex to split the sentences using only the single commas from the CSV format. Once the sentences were split, the double commas could be converted to single commas trivially. The evaluations were somewhat difficult to code, but once their respective formulas were found implementing them was relatively trivial. Unfortunately, due to how SICK-E's classification works, SemanticBERT + cosine similarity is unable to properly classify the information, and a score could not be found.

Unfortunately, the implementation of XLNet was not so simple. The original plan was to use the implementation linked in the original paper [16], but after much consideration the plan was changed to use hugging face transformers [17] instead, as it would be significantly easier to import and use. Initially, there were major issues with the XLNet model, as the input dimensions would consistently be off. This was solved through usage of padding. Then, training and evaluation became an issue, as running either would consistently fail from using too much GPU memory. This was solved by reducing batch steps. After this, I was able to find the accuracy of classification datasets without issue, but unfortunately, finding an XLNet model capable of regression was still a major problem.

Implementing MT-DNN was a major issue from the very beginning. It took several days to figure out atensorflow import that kept going wrong required an older version of Python (3.6), and another day after that to realize the GitHub project used [18] was designed for a Linux OS. After those two problems were found, the location of implementation was switched to a cloud computing server known as Paperspace Gradient. Once accomplished, the model was able to be set up. Import errors made it difficult at times but installing them via apt install worked nicely to clear up the problems. Rewriting code in the program was often necessary to account for running python programs with python3 instead of python2, as well. Furthermore, for the MT-DNN to work quickly, reinstalling a CUDA driver that was only partially installed became necessary, which was difficult as simply deleting it did not work as there were dependencies involved. Ultimately, manually deleting the system programs that relied on said driver, then deleting the driver solved the issue. Uploading the datasets seemed simple at first, but quickly became difficult, as each task had a specific data orientation that the uploaded datasets did not match. After preprocessing the datasets again to account for this and splitting them up into train/test/dev splits, evaluation was thankfully simple.

Implementing the knowledge-based measure was for the most part simple, although there were some hiccups along the way. The initial plan was to use the implementation mentioned in [8], but it was not linked anywhere in the paper and it could not be found online. Ultimately, the implementation was changed to include Sematch [19,20] as the implementation for the knowledge-based measure. Originally, the plan was to combine multiple knowledge-based measures together, but it quickly became clear that doing so would be difficult to impossible in the time remaining. Ultimately, the knowledge-based measure chosen was the Lin method, as it seemed the most effective of the individual algorithms mentioned in [8] and had an easy normalization of between 0 and 1. Having chosen the algorithm and GitHub implementation, the most difficult problem of implementation made itself known. Knowledge-based measures work by turning words into concepts and comparing those concepts, and thus they are designed for word similarity instead of sentence similarity. All the evaluation metrics compared sentence similarity, so it was necessary to convert word similarity to sentence similarity. Thankfully, several papers that did just that were found [12,13]. They achieved this feat by building a matrix of comparisons between each word in the two sentences. Once this was done, the papers took the highest value in the matrix, added it to a list, deleted the row and column the value was in, and continued until the highest value was zero or until there were no columns or rows left. The authors also combined the knowledge-based measure with a similarity measure so that the algorithm would be more sensitive to lexical information. It was trivial to use those papers to convert the implementations' word similarity to sentence similarity, and from there things proceeded smoothly.

Table 3. Result analysis per task/evaluation combination.

|  | **SICK-R Pearson** | **SICK-R Spearman** | **SICK-E Accuracy** | **AFS Pearson** | **AFS Spearman** | **MRPC Accuracy** |
|---|---|---|---|---|---|---|
| **Sentence-BERT + Cosine Similarity** | 73.61% | 75.038% | N/A | 28.034% | 75.063% | 68.017% |
| **MT-DNN** | 92.185% | 88.567% | 90.706% | 80.860% | 78.178% | 88.286% |
| **XLNet2 (second implementation)** | 47.697% | N/A | 57.383% | 72.752% | N/A | 66.236% |
| **XLNet1 (first implementation)** | N/A | N/A | 56.911% | N/A | N/A | 66.983% |
| **Lin Measure + String Similarity** | 60.546% | 75.038% | N/A | 32.273% | 75.063% | 70.172% |

During the XLNet implementation, it became clear that the transformers XLNet code being working with was designed for classification problems and could thus not handle continuous data. As such, it was decided to implement a different implementation linked in the original XLNet paper [16] as well to ensure that XLNet data concerning SICK-R and AFS could be gathered. The main issue was the matter of installing CUDA 10.0, as Paperspace Gradient automatically installed CUDA 10.2, which needed to be uninstalled before CUDA 10.0 was installed. There were also issues with the Nvidia driver, which would prevent CUDA from fully installing. In the end, the issue was resolved by removing all traces of CUDA from my system, partially installing CUDA 10.0, manually removing the nVidia driver and its dependencies, and then running apt --fix-broken install to properly install CUDA 10.0. After this, implementation largely went smoothly. The only real roadblocks after this point was that XLNet only used Pearson similarity as a metric, and it did not naturally include MRPC processing. Coding in an MRPC Processor subclass using their basic MNLI Matched Processor became necessary, after which the evaluation went smoothly. Unfortunately, adding Spearman Rank Similarity was a much more arduous task, and there was no time left to finish the job. All results used 60% train, 20% test and 20% dev datasets. Figure 2 shows the accuracy of document similarity algorithms.

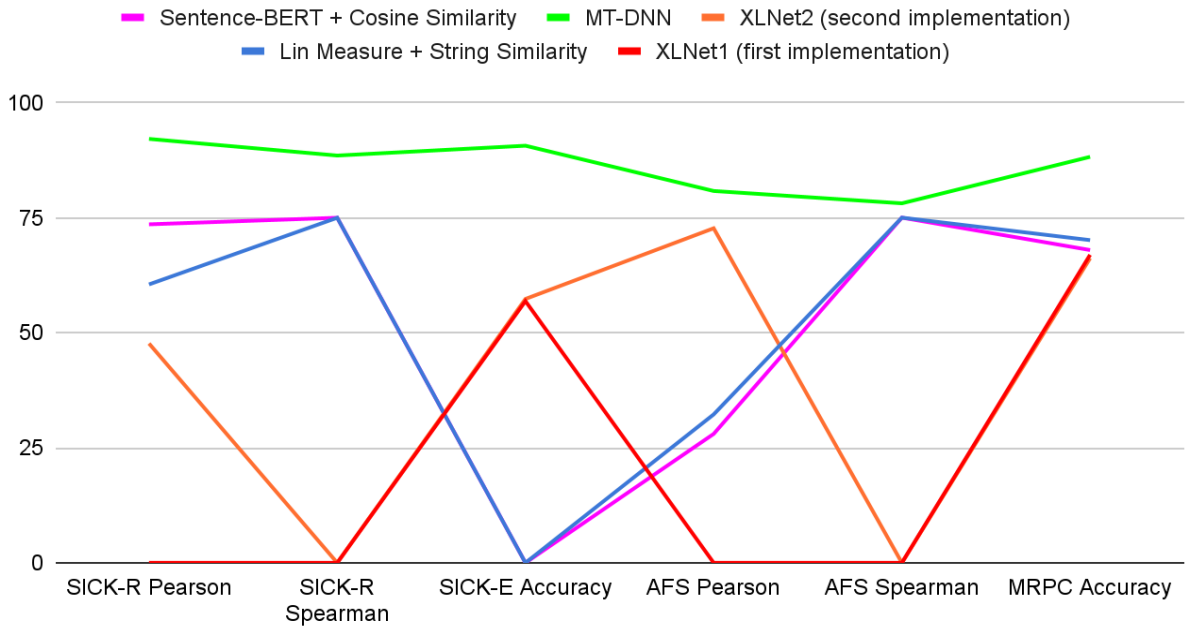

Figure 2. Result analysis per task/evaluation combination.

Both Sentence-Bert + cosine similarity and the unofficial XLNet code was implemented on an Alienware m15 R4 with an Intel i7-10870H CPU, 32 GB of RAM, and a NVIDIA GForce RTX 3070 GPU. The MT-DNN, second XLNet implementation and Lin measure + string similarity was implemented on Paperspace, with an Intel Xeon E5-2623 v4 CPU, 30 GB of RAM, and a NVIDIA P5000 GPU.

Of the algorithms tested, the MT-DNN is by far the most effective, as it was the only algorithm to successfully process all evaluation metrics, and it also had the highest scores in all evaluation metrics. Lin Measure + String Similarity and Semantic BERT + Cosine Similarity had roughly the same results, with Semantic BERT + Cosine Similarity performing better in the SICK-R Pearson task, but slightly worse in the AFS Pearson and MRPC tasks. The official XLNet code performed incredibly well in the AFS Pearson task but performed badly in the SICK-R Pearson task and had a slightly worse MRPC score than Lin + String Similarity and BERT + cosine similarity's scores. However, unlike Lin + String similarity and BERT + Cosine Similarity, the second XLNet implementation was able to give a score for SICK-E, so it has a major advantage in flexibility. The first XLNet implementation had similar scores to the second XLNet implementation, but was unable to handle continuous data, so in terms of versatility and results it performed the worst of all tested algorithms.

In the future, XLNet could be further investigated. The second implementation of XLNet is theoretically capable of using Spearman Rank similarity as a metric, making it the only algorithm other than MT-DNN that can process all evaluation metrics.

## 5. CONCLUSIONS

This research gave a comprehensive study of four algorithms, Semantic BERT + Cosine Similarity, MT-DNN, XLNet, and Lin Measure + String Similarity. Using the MRPC, AFS, SICK-E and SICK-R datasets, and the Spearman Rank, Pearson, and Accuracy evaluation metrics, the research was able to look at each of the four algorithms strengths and weaknesses, and thus give an answer to the question of which document similarity algorithm is most effective.

From the data obtained in this research, the MT-DNN is clearly the most effective document similarity algorithm tested.

## ACRONYMS

MRPC: Microsoft Research Paraphrase Corpus.
AFS: Argument Facet Similarity.
SICK: Sentences Involving Composition Knowledge.
BERT: Bidirectional Encoder Representations from Transformers.
MT-DNN: Multi-Task Deep Neural Network.